\documentclass[10pt]{article}
\usepackage{amsmath}
\usepackage{amssymb}
\usepackage{booktabs}
\usepackage[dvipsnames]{xcolor}
\usepackage[most]{tcolorbox}

\usepackage[round]{natbib}
\usepackage{geometry}
\usepackage{graphicx}

\usepackage{charter}
\usepackage{hyperref}
\hypersetup{colorlinks=true,allcolors=BlueViolet}

\title{\textsc{\textbf{Facilitating Advanced Sentinel-2 Analysis Through a Simplified Computation of\\Nadir BRDF Adjusted Reflectance}}}

\author{
David Montero\textsuperscript{1,2,*}\quad
Miguel D. Mahecha\textsuperscript{1,2,3}\quad
César Aybar\textsuperscript{4}\quad
Clemens Mosig\textsuperscript{1,3}\\
Sebastian Wieneke\textsuperscript{1}\\
\small\textsuperscript{1}RSC4Earth, IEF, Leipzig University\quad
\textsuperscript{2}iDiv\quad \textsuperscript{3}ScaDS.AI\quad \textsuperscript{4}IPL, Universitat de Val\`encia\\
\small*Corresponding author: \texttt{david.montero@uni-leipzig.de}
}
\date{}

\begin{document}

\maketitle

\begin{abstract}
The Sentinel-2 (S2) mission from the European Space Agency's Copernicus program provides essential data for Earth surface analysis. Its Level-2A products deliver high-to-medium resolution (10-60 m) surface reflectance (SR) data through the MultiSpectral Instrument (MSI). To enhance the accuracy and comparability of SR data, adjustments simulating a nadir viewing perspective are essential. These corrections address the anisotropic nature of SR and the variability in sun and observation angles, ensuring consistent image comparisons over time and under different conditions. The $c$-factor method, a simple yet effective algorithm, adjusts observed S2 SR by using the MODIS BRDF model to achieve Nadir BRDF Adjusted Reflectance (NBAR). Despite the straightforward application of the $c$-factor to individual images, a cohesive Python framework for its application across multiple S2 images and Earth System Data Cubes (ESDCs) from cloud-stored data has been lacking. Here we introduce \texttt{sen2nbar}, a Python package crafted to convert S2 SR data to NBAR, supporting both individual images and ESDCs derived from cloud-stored data. This package simplifies the conversion of S2 SR data to NBAR via a single function, organized into modules for efficient process management. By facilitating NBAR conversion for both SAFE files and ESDCs from SpatioTemporal Asset Catalogs (STAC), \texttt{sen2nbar} is developed as a flexible tool that can handle diverse data format requirements. We anticipate that \texttt{sen2nbar} will considerably contribute to the standardization and harmonization of S2 data, offering a robust solution for a diverse range of users across various applications. \texttt{sen2nbar} is an open-source tool available at \url{https://github.com/ESDS-Leipzig/sen2nbar}.
\end{abstract}

\section{Introduction}

The Sentinel-2 (S2) mission from the Copernicus program of the European Space Agency (ESA), comprising the Sentinel-2A (S2A) and Sentinel-2B (S2B) satellites, is equipped with the MultiSpectral Instrument (MSI) sensor \citep{drusch2012sentinel2}. This sensor is designed to capture data across various spectral bands with high-to-medium spatial resolutions (Table~\ref{tab:spectral-parameters}). The Level-2A product from S2 provides surface reflectance (SR) data, which are indispensable for detailed Earth surface analysis. This product has been employed in diverse applications, examples include the estimation of carbon fluxes \citep{pabon2022potential_s2_gpp}, the investigation of vegetation dynamics using spectral indices \citep{montero2023asi}, and Land Use and Land Cover (LULC) products generation via Artificial Intelligence (AI) models \citep{brown2022dynamic_world}. The utility of S2 SR data is significantly enhanced through the incorporation into Earth System Data Cubes \citep[ESDCs, ][]{mahecha2020esdc, montero2023datacubes_challenges}, which offer an organized spatiotemporal framework, allowing for simplified multitemporal analyses.

\begin{table*}[h]
\footnotesize
\centering
\begin{tabular}{ l l c c c c c c c }\toprule
\textbf{Name} & \textbf{Band} & \textbf{Res. (m)} & \multicolumn{2}{c}{\textbf{Wavelength (nm)}} & $f_{\text{iso}}$ & $f_{\text{geo}}$ & $f_{\text{vol}}$ & \textbf{Reference} \\\cline{4-5}
     &      &          & \textbf{Sentinel-2A} & \textbf{Sentinel-2B} &  &  &  &  \\\toprule
Blue          & B02 & 10 & 496.6 & 492.1 & 0.0774 & 0.0079 & 0.0372 & \cite{roy2017sentinel2examination} \\
Green         & B03 & 10 & 560.0 & 559.0 & 0.1306 & 0.0178 & 0.0580 & \cite{roy2017sentinel2examination} \\
Red           & B04 & 10 & 664.5 & 665.0 & 0.1690 & 0.0227 & 0.0574 & \cite{roy2017sentinel2examination} \\
Red Edge 1    & B05 & 20 & 703.9 & 703.8 & 0.2085 & 0.0256 & 0.0845 & \cite{roy2017rededge} \\
Red Edge 2    & B06 & 20 & 740.2 & 739.1 & 0.2316 & 0.0273 & 0.1003 & \cite{roy2017rededge} \\
Red Edge 3    & B07 & 20 & 782.5 & 779.7 & 0.2599 & 0.0294 & 0.1197 & \cite{roy2017rededge} \\
NIR           & B08 & 10 & 835.1 & 833.0 & 0.3093 & 0.0330 & 0.1535 & \cite{roy2017sentinel2examination} \\
SWIR 1        & B11 & 20 & 1613.7 & 1610.4 & 0.3430 & 0.0453 & 0.1154 & \cite{roy2017sentinel2examination} \\
SWIR 2        & B12 & 20 & 2202.4 & 2185.7 & 0.2658 & 0.0387 & 0.0639 & \cite{roy2017sentinel2examination} \\\bottomrule
\end{tabular}
\caption{BRDF spectral parameters for the Sentinel-2 bands.}
\label{tab:spectral-parameters}
\end{table*}

While S2 SR data is widely used, its viewing angle ($\pm10.3$°) and field of view (20.6°) amplify Bidirectional Reflectance Distribution Function (BRDF) effects due to SR anisotropy \citep{roy2017sentinel2examination}. To minimize BRDF effects, adjustments simulating a nadir viewing perspective are needed \citep{roy2016landsat_nbar}. These corrections address the directional effects arising from the anisotropic nature of SR and the variability of sunlight and satellite viewing angles. Such adjustments are crucial for the consistent comparison of images taken at different times and under various sensor acquisition conditions. This is particularly important for the processing and analysis of analysis-ready ESDCs, which are increasingly utilized due to their organized spatiotemporal structure and the simplicity of generating them from cloud-stored data \citep{montero2023datacubes_challenges}.

The spectral parameters derived from the MODIS BRDF model facilitate the computation of directional reflectance across any specified sensor viewing and solar angles \citep{roy2008modis_landsat}. Using this framework, \cite{roy2008modis_landsat,roy2016landsat_nbar} introduced the $c$-factor method, a straightforward approach for adjusting Landsat SR data by applying the MODIS spectral BRDF model parameters. This adjustment yields Nadir BRDF Adjusted Reflectance (NBAR) by multiplying the observed Landsat SR with the ratio of reflectances predicted by the MODIS BRDF model for both the observed Landsat SR and a standard nadir view under fixed solar zenith conditions. \citet{roy2017sentinel2examination} and \citet{roy2017rededge} extended this methodology for multiple S2 spectral bands. Despite the simplicity of the $c$-factor method for individual S2 images, a unified Python framework for applying this conversion uniformly across multiple images, especially for ESDCs derived from SpatioTemporal Asset Catalogs (STAC), is missing so far.

This paper presents \texttt{sen2nbar}, an open-source Python package designed to facilitate converting S2 SR data to NBAR using the $c$-factor method. \texttt{sen2nbar} is engineered to support both automatic NBAR conversions through a single function, accommodating SAFE files and ESDCs derived from STAC. The document is organized into the following sections: Section~\ref{sec:package} presents the \texttt{sen2nbar} framework, elaborating on the modules that simplify the processing steps; Section~\ref{sec:showcase} illustrates the application of \texttt{sen2nbar}, providing practical examples; Section~\ref{sec:discussion} explores the limitations of \texttt{sen2nbar} as well as its potential, with a particular emphasis on AI; and Section~\ref{sec:conclusions} summarizes our conclusions.

\section{The \texttt{sen2nbar} Python package}
\label{sec:package}

The \texttt{sen2nbar} package, developed in Python, facilitates the conversion of S2 L2A Surface Reflectance values into NBAR values through a single function. To achieve this higher level, the package is structured into several modules, ensuring a methodical conversion process leveraging multidimensional arrays via \texttt{xarray} \citep{hoyer2017xarray} and \texttt{numpy} \citep{harris2020numpy}:

\begin{figure*}[ht!]
\begin{center}
    \includegraphics[width=1.0\textwidth]{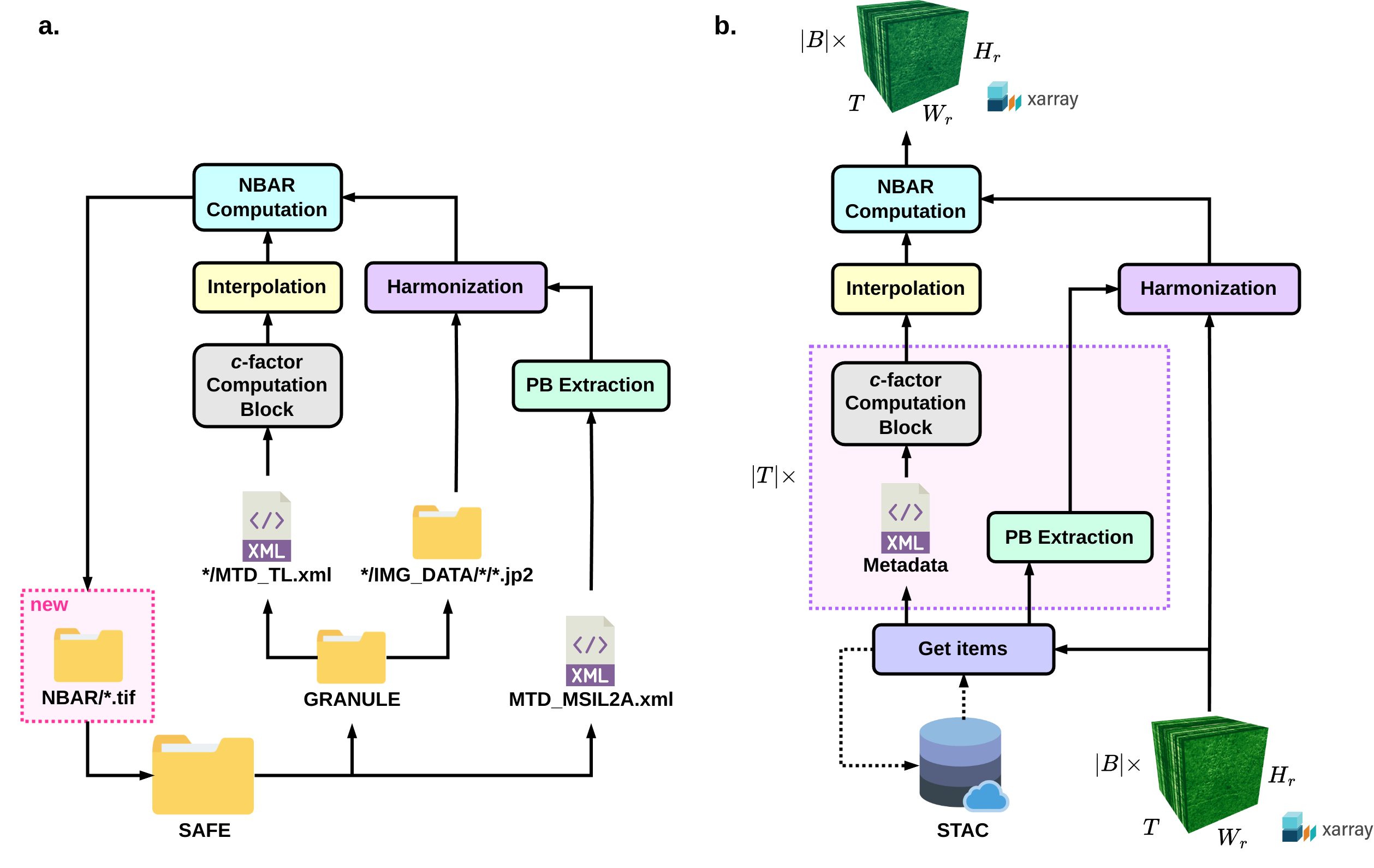}
    \caption{Flowcharts depicting the process of computing NBAR for SAFE files (a) and ESDCs (b). In (a), NBAR values are calculated for each spectral band and subsequently stored either as GeoTIFF or Cloud Optimized GeoTIFF (COG) within a newly created folder titled ``NBAR'' inside the original SAFE file's directory. For ESDCs, depicted in (b), the NBAR computation results in the generation of a new ESDC, represented as an \texttt{xarray} object. The dashed arrows linking the STAC object in (b) signify that this procedure is automated, eliminating the need for users to manually extract items from the STAC.}
\label{fig:nbar}
\end{center}
\end{figure*}

\subsection{The \texttt{axioms} module}

This module stores the predefined MODIS BRDF spectral model parameters, namely $f_{\text{iso}}(\lambda)$, $f_{\text{vol}}(\lambda)$, and $f_{\text{geo}}(\lambda)$ (as shown in Table~\ref{tab:spectral-parameters}), in addition to preserving the values corresponding to the spatial resolution of each spectral band. In accordance to the methodology proposed by \citet{roy2017rededge,roy2017sentinel2examination}, our implementation encompasses all spectral bands except for the Aerosols (B01) and the narrow NIR (B8A) bands. Consequently, the NBAR values are not computed for B01 and B8A bands. Also note that for the S2 Red Edge bands, the BRDF spectral model parameters were determined through a process of linear interpolation between the red and NIR MODIS BRDF spectral model parameters \citep{roy2017rededge}.

\subsection{The \texttt{metadata} module}
\label{sec:metadata}

This module is tasked with the extraction of two critical sets of data from the scene's metadata files (\texttt{MTD\_MSIL2A.xml} and \texttt{MTD\_TL.xml}, Figure~\ref{fig:nbar}a), essential for the computation and harmonization of NBAR values: firstly, it extracts the sun and sensor viewing angles; secondly, it determines the processing baseline of the scene. The acquisition of this information is conducted directly from the metadata file when dealing with SAFE files\footnote{\url{https://sentinels.copernicus.eu/web/sentinel/user-guides/sentinel-2-msi/data-formats}}. Alternatively, when operating with ESDCs constructed via \texttt{stackstac} or \texttt{cubo} \citep{montero2024ondemand}, the module retrieves the requisite data from the corresponding metadata STAC asset within a specific item in a STAC catalog using the \texttt{requests} library. This dual-path approach ensures flexibility and adaptability in processing various data sources for NBAR computation and harmonization.

\subsubsection{Solar and sensor viewing angles:}

The extraction process for solar and sensor viewing angles in degrees involves the aggregation of detector-specific data across each spectral band into a unified 4-dimensional array $\mathbf{A}\in\mathbb{R}^{|\Lambda|\times|\Theta|\times H \times W}$. This array is structured with dimensions $|\Lambda|=|B|+1$, where $B$ are the bands undergoing conversion (with an additional index allocated for solar information) and $\Lambda$ is the set of bands added to the solar component; $|\Theta|=2$ (the number of extracted angles, i.e. $\Theta=\{\text{zenith, azimuth}\}$); and $H=W=23$, reflecting the spatial resolution of this information (5 km). The assignment of spatial coordinate values within this grid uses the Upper Left (UL) X (ULX) and Y (ULY) coordinates, derived from the tile's geocoding information.

\subsubsection{Processing Baseline:}

The Processing Baseline (PB) of the scene is extracted and converted into a floating-point numerical format. This PB version serves as a criterion for initiating a harmonization procedure for scenes possessing a PB version of 4.00 or greater.

\subsection{The \texttt{kernels} module}
\label{sec:kernels}

A kernel-based BRDF model integrates various scattering mechanisms as a linear combination of distinct scattering modes, termed kernels \citep{lucht2000albedo_brdf_model}. The Ross-Li BRDF model (Section~\ref{sec:brdf}), also known as the Ross-Thick/Li-Sparse Reciprocal BRDF model, which is the standard for MODIS BRDF \citep{lucht2000albedo_brdf_model}, is composed of three components: the isotropic scattering parameter, the radiative transfer-type volumetric scattering kernel (Volumetric Kernel $K_{\text{vol}}$), and geometric-optical surface scattering kernel (Geometric Kernel $K_{\text{geo}}$). The objective of this module is to calculate the volumetric and geometric kernels. The computation of these kernels adheres to the mathematical formulations presented by \cite{lucht2000albedo_brdf_model}. Here we use the solar zenith angle $\theta=\mathbf{A}_{\{\text{sun, zenith}\}}$, the view zenith angle $\vartheta=\mathbf{A}_{\{B\text{, zenith}\}}$, and the view-sun relative azimuth angle $\phi=\mathbf{A}_{\{\text{sun, azimuth}\}}-\mathbf{A}_{\{B\text{, azimuth}\}}$ (all angles expressed in radians). Note that $\theta\in\mathbb{R}^{H\times W}$ (since it is a 2-dimensional array representing the solar zenith angle) while $\vartheta,\phi\in\mathbb{R}^{|B|\times H\times W}$ (since they are 3-dimensional arrays representing the zenith and relative azimuth angles for all bands). This implies that $K_{\text{vol}}(\theta,\vartheta,\phi)$, $K_{\text{geo}}(\theta,\vartheta,\phi)\in\mathbb{R}^{|B|\times H\times W}$:

\subsubsection{Volumetric Kernel:}

The Volumetric Kernel $K_{\text{vol}}$ in the Ross-Li BRDF model is represented by the RossThick kernel \citep{roujean1992kvol}, which is based on the radiative transfer theory presented by \cite{ross1981radiation_plant_stands}:

\begin{equation}
    K_{\text{vol}}(\theta,\vartheta,\phi) = \frac{(\pi/2 - \xi) \cos \xi + \sin \xi}{\cos \theta + \cos \vartheta} - \frac{\pi}{4}
\end{equation}

Where $\cos \xi = \cos \theta \cos \vartheta + \sin \theta \sin \vartheta \cos \phi$.

\subsubsection{Geometric Kernel:}

The Geometric Kernel $K_{\text{geo}}$ in the Ross-Li BRDF model is represented by the LiSparse kernel \citep{wanner1995kgeo}, which is based on the geometric-optical mutual shadowing BRDF model by \cite{li1992geo_optical_brdf}:
\begin{equation}
        K_{\text{geo}}(\theta,\vartheta,\phi) = O - \sec \theta' - \sec \vartheta' + \frac{1}{2} (1 + \cos \xi') \sec \theta' \sec \vartheta'
\end{equation}
where
\begin{align}
  O &= \frac{1}{\pi} (t - \sin t \cos t) (\sec \theta' + \sec \vartheta') \\
  \cos t &= \frac{h}{b} \frac{\sqrt{D^2 + (\tan \theta' \tan \vartheta' \sin \phi)^2}}{\sec \theta' + \sec \vartheta'} \\
  D &= \sqrt{\tan^2 \theta' + \tan^2 \vartheta' - 2 \tan \theta' \tan \vartheta' \cos \phi} \\
  \cos \xi' &= \cos \theta' \cos \vartheta' + \sin \theta' \sin \vartheta' \cos \phi \\
  \theta' &= \tan^{-1} \left( \frac{b}{r} \tan \theta \right) \\
  \vartheta' &= \tan^{-1} \left( \frac{b}{r} \tan \vartheta \right)
\end{align}
Here, the $\cos t$ term is constrained to [-1,1], $h/b=~2$, and $b/r=~1$ \citep{lucht2000albedo_brdf_model}.

\subsection{The \texttt{brdf} module}
\label{sec:brdf}

This module is dedicated to the computation of the Ross-Li BRDF model, leveraging the scattering kernels sourced from the \texttt{kernels} module and the BRDF spectral parameters obtained from the \texttt{axioms} module, in accordance with the mathematical framework outlined by \cite{lucht2000albedo_brdf_model}. Prior to computation, the BRDF spectral model parameters are transformed into \texttt{xarray.DataArray} objects such that $f_{\text{iso}},$ $f_{\text{vol}},$ $f_{\text{geo}}$ $\in\mathbb{R}^{|B|}$. This enhances the efficiency and scalability of the BRDF model computation. Consequently, this computation of the Ross-Li BRDF model outputs values as a multidimensional array $\text{BRDF}(\theta,\vartheta,\phi)\in\mathbb{R}^{|B|\times H\times W}$:

\begin{equation}
        \text{BRDF}(\theta,\vartheta,\phi) = f_{\text{iso}} + f_{\text{vol}}K_{\text{vol}}(\theta,\vartheta,\phi) + f_{\text{geo}}K_{\text{geo}}(\theta,\vartheta,\phi)
\end{equation}

\subsection{The \texttt{c\_factor} module}
\label{sec:cfactor}

This module is tasked with calculating the  $c$-factor, adhering to the methodology delineated by \cite{roy2008modis_landsat}. The $c$-factor is instrumental in modifying the reflectance values to align with any given viewing or solar geometry, as outlined by \cite{roy2016landsat_nbar}. To standardize the reflectance measurements to a nadir viewing zenith angle, the value of the viewing zenith angle, $\vartheta$, is predetermined to be 0. Note that $c(\theta,\vartheta,\phi)\in\mathbb{R}^{|B|\times H\times W}$:

\begin{equation}
    c(\theta,\vartheta,\phi) = \frac{\text{BRDF}(\theta,0,\phi)}{\text{BRDF}(\theta,\vartheta,\phi)}  
\end{equation}

Furthermore, this module integrates information and functionalities from preceding modules to directly derive the $c$-factor from a metadata file, establishing a comprehensive function specifically tailored for SAFE files. Additionally, recognizing the accessibility of metadata files through STAC items, a separate function has been developed to compute the $c$-factor directly from STAC items. This latter function is notably equipped with the capability to reproject the $c$-factor to a designated CRS when required. This reprojection feature is essential for ensuring the alignment and consistency of ESDCs created from scenes that originate in varying CRS, thereby facilitating their integration and analysis within a unified spatial framework.

\begin{figure*}[ht!]
\begin{center}
    \includegraphics[width=1.0\textwidth]{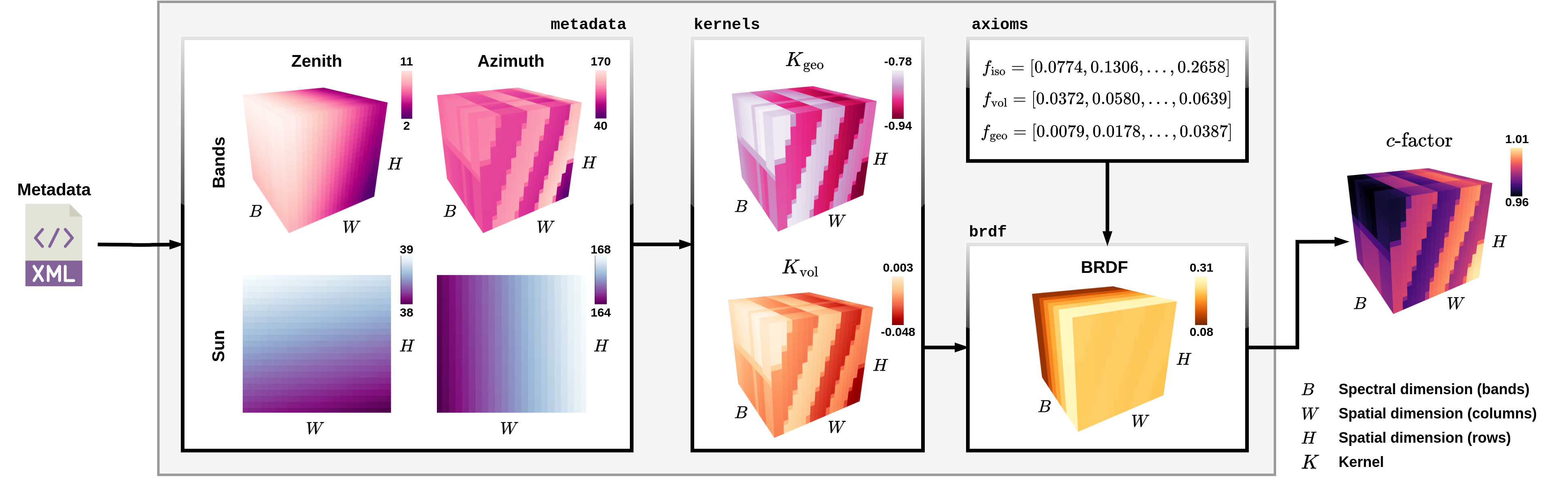}
    \caption{Diagram illustrating the $c$-factor computation block. The arrays depicted in this block serve as examples derived from a single metadata file, with the understanding that values could vary across different files.}
\label{fig:cfactor}
\end{center}
\end{figure*}

\subsection{The \texttt{nbar} module}

This module is designed to compute NBAR values, tailored to the format of the input data, whether they are SAFE files or ESDCs constructed via \texttt{stackstac} or \texttt{cubo}:

\subsubsection{SAFE files:}

For SAFE files (Figure~\ref{fig:nbar}a), the spatial resolution is band-specific, denoted as $\lambda\in B$, leading to the necessity of computing NBAR values for each $\lambda$ independently. Let $\rho_{\lambda}\in\mathbb{R}^{H_{\lambda}\times W_{\lambda}}$ represent the SR of band $\lambda$, with $H_{\lambda}$ and $W_{\lambda}$ indicating its spatial dimensions. Initially, metadata files are used to extract the PB $\in\mathbb{R}_+$ and to compute the $c$-factor $c(\theta,\vartheta,\phi)\in\mathbb{R}^{|B|\times H\times W}$. The PB is then applied to harmonize $\rho_{\lambda}$ if required:

\begin{equation}
    \rho*_{\lambda} =
    \begin{cases} 
    \rho_{\lambda} - 1000, & \text{if PB} \geq4, \\
    \rho_{\lambda}, & \text{otherwise}
    \end{cases}
\end{equation}

Here, $\rho*_{\lambda}$ signifies the harmonized SR values. Subsequently, the NBAR values are computed. It's important to note that the $c$-factor is derived using the original spatial resolution of the angle information (5 km), which makes the computation faster compared to performing angle interpolation beforehand. Therefore, $c(\theta,\vartheta_{\lambda},\phi_{\lambda})$ undergoes bilinear interpolation, denoted by the function $I$ such that $I:\mathbb{R}^{H\times W}\rightarrow \mathbb{R}^{H_{\lambda}\times W_{\lambda}}$, to align with the original resolution of band $\lambda$. The NBAR values are then generated as follows:

\begin{equation}
    \text{NBAR}_{\lambda} = I(c(\theta,\vartheta_{\lambda},\phi_{\lambda}))\times\rho*_{\lambda}  
\end{equation}

Resulting NBAR images are subsequently exported individually into the same SAFE file directory, available in either COG or GeoTIFF formats.

\subsubsection{Earth System Data Cubes:}

For ESDCs constructed from STAC (Figure~\ref{fig:nbar}b), the spatial resolution is predefined, enabling the generation of NBAR values for the entire ESDC via operations on multidimensional arrays. We denoted $\rho_T\in\mathbb{R}^{|T|\times |B|\times H_r\times W_r}$ as the SR of the ESDC, where $T$ represents the temporal dimension items, and $H_r$ and $W_r$ are its spatial dimensions at resolution $r$, alongside the bounding box (BBox) of the ESDC. Initially, the module accesses the STAC, retrieving metadata files and PB values for each item $\tau\in T$. For each $\tau$, the $c$-factor $c_{\tau}(\theta_{\tau},\vartheta_{\tau},\phi_{\tau})$ is calculated and subsequently concatenated along the temporal dimension such that $c_T=\left[c_{\tau}(\theta_{\tau},\vartheta_{\tau},\phi_{\tau})\right]_{{\tau}\in T}$, resulting in $c_T\in\mathbb{R}^{|T|\times |B|\times H\times W}$. The PB values for each $\tau$ are also concatenated as PB$_T\in\mathbb{R}^{|T|}$. This array facilitates the harmonization of $\rho_T$:

\begin{equation}
    \rho*_T = \left[
    \begin{cases} 
    \rho_{\tau} - 1000, & \text{if PB}_{\tau} \geq 4, \\
    \rho_{\tau}, & \text{otherwise}
    \end{cases}
    \right]_{{\tau} \in T}
\end{equation}

Here, $\rho*_T$ represents the harmonized SR values of the ESDC. Following this, NBAR values are calculated. Given that the $c$-factor is derived from the original spatial resolution of angle information, it undergoes bilinear interpolation through function $I$ such that $I:\mathbb{R}^{|T|\times |B|\times H\times W}\rightarrow \mathbb{R}^{|T|\times |B|\times H_r\times W_r}$, aligning with the predetermined resolution and BBox of $\rho*_T$. This process enables NBAR computation for ESDCs with uniform spatial dimensions ($H_r=W_r$, as in cubes generated via \texttt{cubo}) and for those with non-uniform dimensions ($H_r\neq W_r$, as in cubes generated via \texttt{stackstac}). Note that $I$ only interpolates the spatial dimensions $H$ and $W$. The NBAR values are then derived as follows:

\begin{equation}
    \text{NBAR}_T = I(c_T)\times\rho*_T  
\end{equation}

\section{Showcase}
\label{sec:showcase}

In a case study designed to showcase the capabilities of the \texttt{sen2nbar} package in handling multidimensional data, we utilized an ESDC of S2 L2A data generated for the Hainich Natural National Park area, proximate to the DE-Hai Eddy Covariance tower \citep{knohl2003hainich}, a deciduous broadleaf forest in Germany, using \texttt{cubo} \citep{montero2024ondemand}. The ESDC was crafted with a uniform resolution of 10 meters ($r=10$) across all bands, featuring an edge size of 128 pixels, which translates to $H_r=W_r=128$. This ESDC covers the period from 2021-01-01 to 2023-12-31, focusing on images with less than 10\% cloud coverage for analysis, encompassing a total of 90 images ($|T|=90$). It includes all S2 bands with the exception of B01 and B8A. Data from both satellites (S2A and S2B) were incorporated.

Utilizing the aforementioned ESDC, NBAR values were computed, followed by the calculation of the difference:

\begin{equation}
    \Delta\rho=\text{NBAR}_T-\rho*_T
\end{equation}

\begin{figure*}[b!]
\begin{center}
    \includegraphics[width=1.0\textwidth]{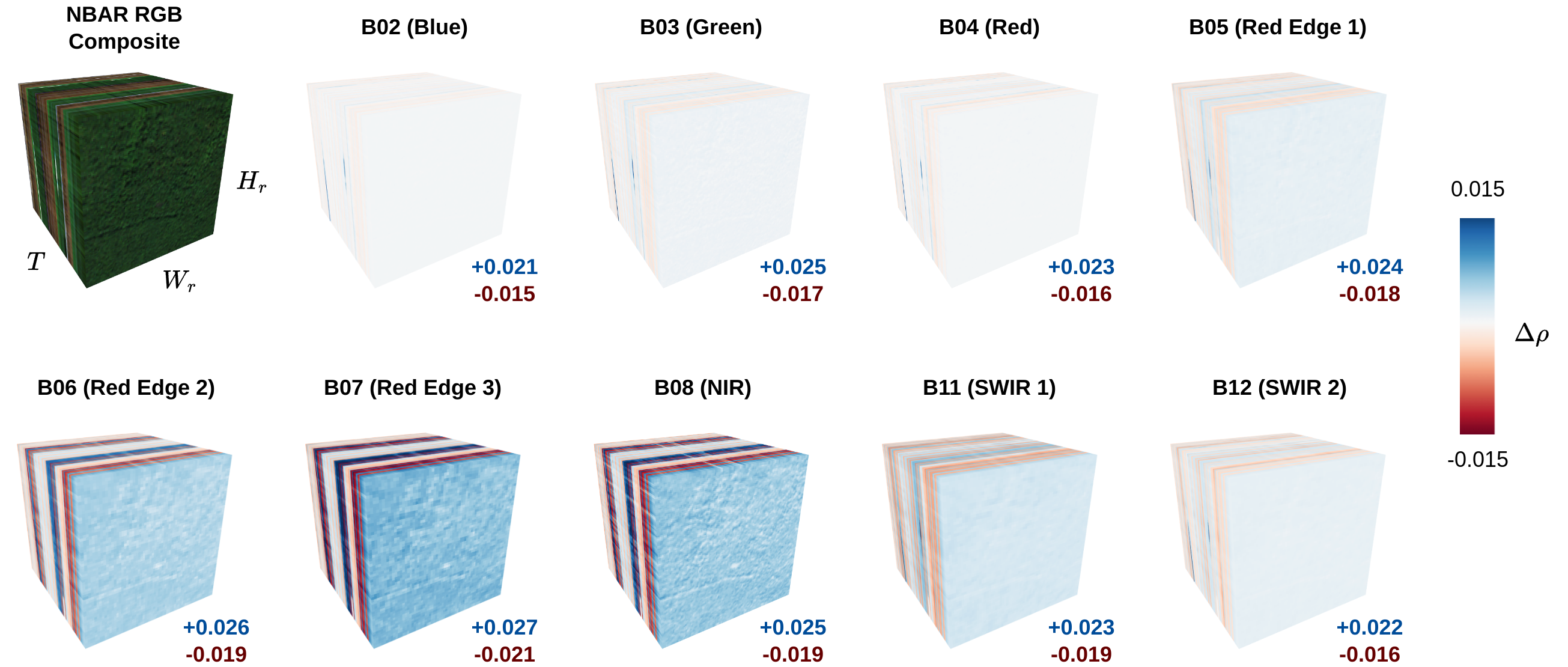}
    \caption{ESDCs illustrating the difference between NBAR and SR values. The initial ESDC presents an RGB composite utilizing visible bands to create a true color reference image. The maximum (blue) and minimum (red) $\Delta\rho$ values for each band are indicated in the lower-right corner of the corresponding ESDC.}
\label{fig:delta-rho}
\end{center}
\end{figure*}

This metric shows the variations in reflectance values when transitioning from SR to NBAR data, particularly within the spatiotemporal framework of the ESDC. Figure~\ref{fig:delta-rho} illustrates the $\Delta\rho_{\lambda}$ for the ESDCs. Spatially, while the reflectance scales according to the $c$-factor and shows minimal variance owing to the confined area of study, the temporal variations are more pronounced, underscoring the significance of NBAR for accurate multitemporal analysis. This effect is especially marked in the Red Edge-NIR region, where discrepancies between values are more substantial compared to other bands. However, it's important to note that the maximum and minimum $\Delta\rho$ values exhibit similar ranges across all bands, highlighting the impact of NBAR adjustments.

Reflectance changes, as highlighted previously, can greatly influence the derivation of various products. To illustrate this effect, we computed four widely used vegetation indices for both the NBAR$_T$ and $\rho*_T$: the Normalized Difference Vegetation Index \citep[NDVI, ][]{rouse1974ndvi}, the Near-Infrared Reflectance of Vegetation \citep[NIRv, ][]{badgley2017nirv}, the Kernel NDVI \citep[kNDVI, ][]{campsvalls2021kndvi}, and the Inverted Red Edge Chlorophyll Index \citep[IRECI, ][]{frampton2013ireci}. Additionally, the difference between the indices derived from NBAR values and those from SR values was calculated as: 

\begin{equation}
    \Delta\psi=\psi_{\text{NBAR}}-\psi_{\rho*}
\end{equation}

\begin{figure*}[ht!]
\begin{center}
    \includegraphics[width=1.0\textwidth]{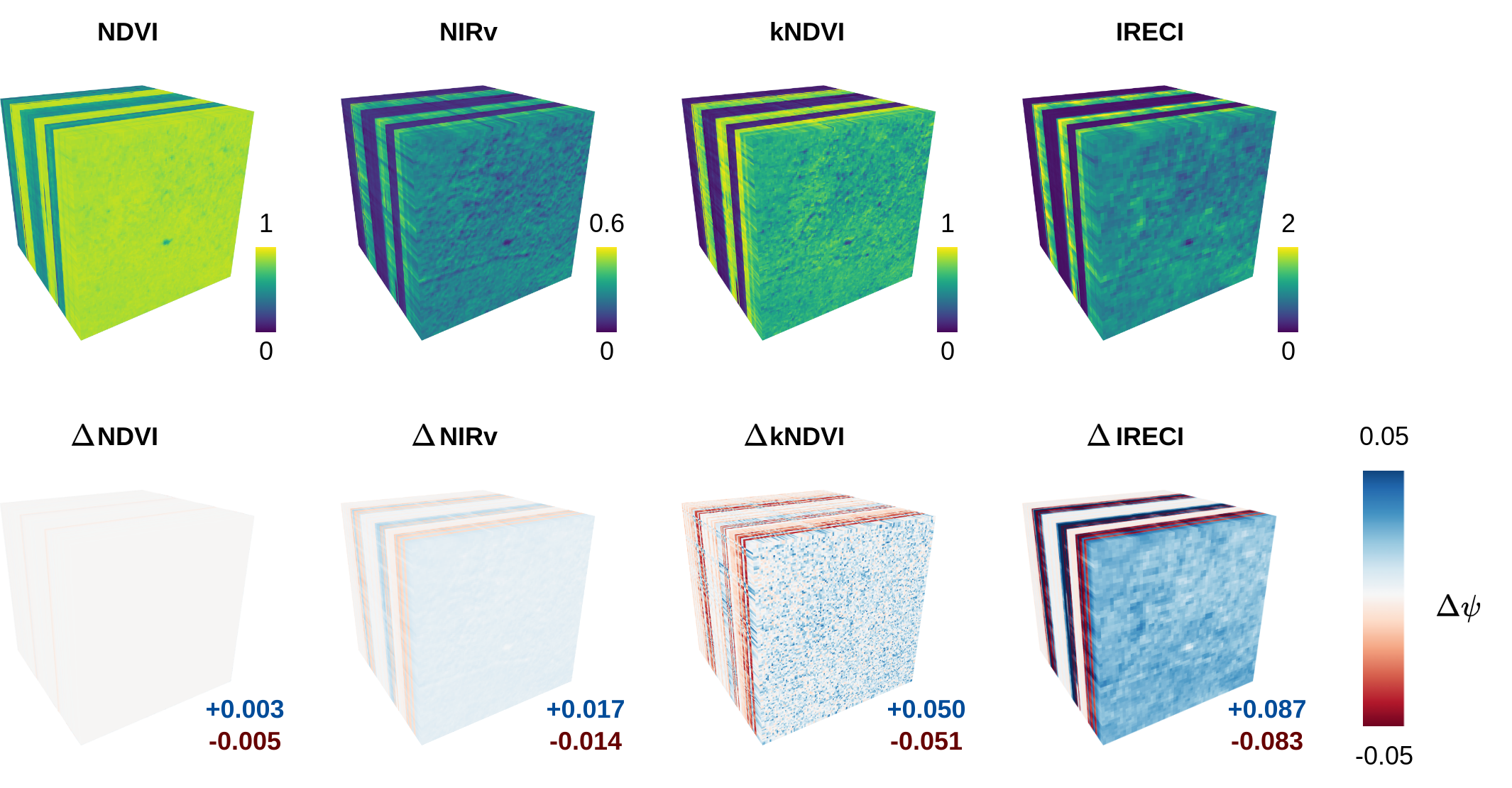}
    \caption{ESDCs illustrating the difference between indices calculated using NBAR and SR values. The first row presents the four calculated indices using the SR values. The second row presents the $\Delta\psi$ values for each ESDC. The maximum (green) and minimum (brown) $\Delta\psi$ values for each index are indicated in the lower-right corner of the corresponding ESDC.}
\label{fig:delta-psi}
\end{center}
\end{figure*}

Here, $\psi$ represents the index in question. Figure~\ref{fig:delta-psi} displays the $\Delta\psi$ for the ESDCs. It is noted that the impact on NDVI is minimal, but it considerably increases for the other indices, particularly for kNDVI and IRECI, where the largest absolute changes observed reach values of 0.051 and 0.087, respectively. This variance underlines that the influence of reflectance adjustments on derived indices is index-dependent and can be substantial enough to potentially skew analysis reliant on these derived features.

\section{Discussion}
\label{sec:discussion}

Here we present various factors that either constrain or enhance the capabilities of \texttt{sen2nbar}, focusing on data providers and scalability. As we discuss below, this approach enhances the possibilities for big data computations, workflow generation towards analysis-ready data cubes, and therefore eases the application of novel artificial intelligence (AI) methods.

\subsection{The Impact of Metadata Availability}

\texttt{sen2nbar} is fully operational with SAFE files, which are widely accessible as they are publicly available (including metadata files). When it comes to ESDCs, \texttt{sen2nbar} is compatible with those generated from STAC, particularly from public collections. A prime example of this compatibility is with the S2 L2A collection from the Microsoft Planetary Computer STAC, which possesses the required metadata files for transforming SR to NBAR values. Conversely, \texttt{sen2nbar} does not support the S2 L2A collection from the Element84 STAC due to the absence of necessary metadata files. Similarly, while tools like \texttt{cubo} facilitate the creation of ESDCs from Google Earth Engine (GEE), the lack of metadata files in this context precludes the calculation of NBAR values. Future versions of \texttt{sen2nbar} may overcome these limitations by enabling the interconnection of STAC assets (leveraging data from one STAC provider while sourcing metadata from another). This approach, however, depends on the availability of identical scenes across different collections.

\subsection{Parallel Processing and Lazy Evaluation}

\texttt{sen2nbar} employs multidimensional arrays through \texttt{xarray} and \texttt{numpy}, enabling the execution of various array operations. In the context of ESDCs, the outcome of NBAR processing is another multidimensional array mirroring the dimensions of the input array. This architecture facilitates the acceleration of computations by exploiting lazy evaluations and parallel processing when necessary. Specifically, when an ESDC is generated via \texttt{stackstac} or \texttt{cubo}, the cube initially remains unevaluated, and its actual creation occurs only upon data retrieval. This setup allows the NBAR processing to be integrated into the lazy evaluation chain, as \texttt{xarray} accommodates \texttt{dask} arrays \citep{rocklin2015dask}, ensuring that the NBAR computation is conducted concurrently with cube creation. Moreover, if an ESDC is pre-existing, either stored in memory or on disk, it can be divided into chunks for processing. \texttt{sen2nbar} is then capable of executing the NBAR transformation in a lazy manner on these chunks. This approach is particularly effective with inherently chunked data formats, such as zarr, which can be processed lazily and re-written to disk in chunks, enhancing efficiency.

\subsection{Refining Multidimensional Dataset Production}

\texttt{sen2nbar} can enhance the accuracy of S2 data and derived features, supporting a wide range of applications, from individual snapshot analyses to comprehensive multitemporal studies. This extends to datasets generated from S2 data. This improvement will be largely attributed to the streamlined generation of ESDCs via STAC coupled with the data accuracy refinement process offered by \texttt{sen2nbar}. This conversion process ensures interoperability across the temporal dimension for all kind of applications. As a result, \texttt{sen2nbar} streamlines the processing of multidimensional datasets like FluxnetEO \citep{walther2022fluxneteo}, which originates from MODIS NBAR data designed for vegetation monitoring but is adaptable to S2 data, or BigEarthNet \citep{sumbul2019bigearthnet}, derived from S2 patches aimed at land cover classification benchmarks. Therefore, the \texttt{sen2nbar} package emerges as a crucial tool in remote sensing, improving the accuracy and utility of data for a broad spectrum of applications.

\subsection{Artificial Intelligence}

The ultimate impact of \texttt{sen2nbar} is its potential to greatly improve AI models in remote sensing. With the introduction of Transformer architectures \citep{vaswani2017transformer}, AI has seen the development of new models that surpass traditional architectures across a variety of tasks, including Convolutional Neural Networks (CNNs) and Long-Short Term Memory (LSTM) networks. In remote sensing, the adoption of Transformer models has notably advanced, especially with pretrained models such as SITS-Former \citep{yuan2022sitsformer}. These foundation models, capable of learning from unlabeled datasets, are instrumental for various specific applications. Notable examples include Prithvi-100M \citep{jakubik2023prithvi100m} and SpectralGPT \citep{hong2023spectralgpt}, which are trained on Harmonized Landsat Sentinel \citep[HLS, ][]{claverie2018hls} data and BigEarthNet, respectively. As these models form the basis for diverse applications, the role of \texttt{sen2nbar} in improving the quality of input datasets becomes invaluable. By ensuring that these models have access to clean and accurate data, particularly in a multitemporal context, \texttt{sen2nbar} will greatly contribute to the improvement of AI models' performance in remote sensing applications.

\section{Conclusions}
\label{sec:conclusions}

In this paper, we introduced \texttt{sen2nbar}, an open-source Python package crafted to facilitate the computation of Nadir BRDF Adjusted Reflectance (NBAR) values for Sentinel-2 L2A data. By enabling the computation of NBAR values through a single function, \texttt{sen2nbar} greatly simplifies the process, demonstrating its adaptability and compatibility with SAFE files as well as Earth System Data Cubes (ESDCs) derived from STAC through \texttt{stackstac} or \texttt{cubo}. We foresee \texttt{sen2nbar} playing an instrumental role in enhancing the accuracy of remote sensing data analyses within Earth system research. Its utility is especially pronounced in spatio-temporal analysis and in strengthening advanced AI-driven tasks via refined Sentinel-2 datasets.

\section*{Acknowledgements}

D.M., M.D.M and S.W. acknowledge support from the ``Digital Forest'' project, Ministry of Lower-Saxony for Science and Culture (MWK) via the program Niedersächsisches Vorab (ZN 3679); and the ``RS4BEF'' project via the iDiv's Flexpool program. D.M. and M.D.M acknowledge support from the ``DeepESDL'' project from the European Space Agency (ESA). C.A. acknowledges support from the National Council of Science, Technology, and Technological Innovation (CONCYTEC, Perú) through the ``PROYECTOS DE INVESTIGACIÓN BÁSICA – 2023-01'' program with contract number PE501083135-2023-PROCIENCIA. M.D.M. and C.M. acknowledge support by the German Aerospace Center, DLR (``ML4Earth'') and by the Federal Ministry of Education and Research of Germany and by Sächsische Staatsministerium für Wissenschaft, Kultur und Tourismus in the programme Center of Excellence for AI-research ``Center for Scalable Data Analytics and Artificial Intelligence Dresden/Leipzig'', project identification number: ScaDS.AI. Arrays within all Figures were visualized utilizing \texttt{lexcube} \citep{soechting2024lexcube}. Figures~\ref{fig:nbar} and \ref{fig:cfactor} have been designed using images from Flaticon.com.

\bibliographystyle{abbrvnat}
\bibliography{Bib}

\begin{thebibliography}{32}
\providecommand{\natexlab}[1]{#1}
\providecommand{\url}[1]{\texttt{#1}}
\expandafter\ifx\csname urlstyle\endcsname\relax
  \providecommand{\doi}[1]{doi: #1}\else
  \providecommand{\doi}{doi: \begingroup \urlstyle{rm}\Url}\fi

\bibitem[Badgley et~al.(2017)Badgley, Field, and Berry]{badgley2017nirv}
G.~Badgley, C.~B. Field, and J.~A. Berry.
\newblock Canopy near-infrared reflectance and terrestrial photosynthesis.
\newblock \emph{Science Advances}, 3\penalty0 (3), Mar. 2017.
\newblock ISSN 2375-2548.
\newblock \doi{10.1126/sciadv.1602244}.
\newblock URL \url{http://dx.doi.org/10.1126/sciadv.1602244}.

\bibitem[Brown et~al.(2022)Brown, Brumby, Guzder-Williams, Birch, Hyde, Mazzariello, Czerwinski, Pasquarella, Haertel, Ilyushchenko, Schwehr, Weisse, Stolle, Hanson, Guinan, Moore, and Tait]{brown2022dynamic_world}
C.~F. Brown, S.~P. Brumby, B.~Guzder-Williams, T.~Birch, S.~B. Hyde, J.~Mazzariello, W.~Czerwinski, V.~J. Pasquarella, R.~Haertel, S.~Ilyushchenko, K.~Schwehr, M.~Weisse, F.~Stolle, C.~Hanson, O.~Guinan, R.~Moore, and A.~M. Tait.
\newblock Dynamic world, near real-time global 10 m land use land cover mapping.
\newblock \emph{Scientific Data}, 9\penalty0 (1), June 2022.
\newblock ISSN 2052-4463.
\newblock \doi{10.1038/s41597-022-01307-4}.
\newblock URL \url{http://dx.doi.org/10.1038/s41597-022-01307-4}.

\bibitem[Camps-Valls et~al.(2021)Camps-Valls, Campos-Taberner, Moreno-Martínez, Walther, Duveiller, Cescatti, Mahecha, Muñoz-Marí, García-Haro, Guanter, Jung, Gamon, Reichstein, and Running]{campsvalls2021kndvi}
G.~Camps-Valls, M.~Campos-Taberner, A.~Moreno-Martínez, S.~Walther, G.~Duveiller, A.~Cescatti, M.~D. Mahecha, J.~Muñoz-Marí, F.~J. García-Haro, L.~Guanter, M.~Jung, J.~A. Gamon, M.~Reichstein, and S.~W. Running.
\newblock A unified vegetation index for quantifying the terrestrial biosphere.
\newblock \emph{Science Advances}, 7\penalty0 (9), Feb. 2021.
\newblock ISSN 2375-2548.
\newblock \doi{10.1126/sciadv.abc7447}.
\newblock URL \url{http://dx.doi.org/10.1126/sciadv.abc7447}.

\bibitem[Claverie et~al.(2018)Claverie, Ju, Masek, Dungan, Vermote, Roger, Skakun, and Justice]{claverie2018hls}
M.~Claverie, J.~Ju, J.~G. Masek, J.~L. Dungan, E.~F. Vermote, J.-C. Roger, S.~V. Skakun, and C.~Justice.
\newblock The harmonized landsat and sentinel-2 surface reflectance data set.
\newblock \emph{Remote Sensing of Environment}, 219:\penalty0 145–161, Dec. 2018.
\newblock ISSN 0034-4257.
\newblock \doi{10.1016/j.rse.2018.09.002}.
\newblock URL \url{http://dx.doi.org/10.1016/j.rse.2018.09.002}.

\bibitem[Drusch et~al.(2012)Drusch, Del~Bello, Carlier, Colin, Fernandez, Gascon, Hoersch, Isola, Laberinti, Martimort, Meygret, Spoto, Sy, Marchese, and Bargellini]{drusch2012sentinel2}
M.~Drusch, U.~Del~Bello, S.~Carlier, O.~Colin, V.~Fernandez, F.~Gascon, B.~Hoersch, C.~Isola, P.~Laberinti, P.~Martimort, A.~Meygret, F.~Spoto, O.~Sy, F.~Marchese, and P.~Bargellini.
\newblock Sentinel-2: Esa’s optical high-resolution mission for gmes operational services.
\newblock \emph{Remote Sensing of Environment}, 120:\penalty0 25–36, May 2012.
\newblock ISSN 0034-4257.
\newblock \doi{10.1016/j.rse.2011.11.026}.
\newblock URL \url{http://dx.doi.org/10.1016/j.rse.2011.11.026}.

\bibitem[Frampton et~al.(2013)Frampton, Dash, Watmough, and Milton]{frampton2013ireci}
W.~J. Frampton, J.~Dash, G.~Watmough, and E.~J. Milton.
\newblock Evaluating the capabilities of sentinel-2 for quantitative estimation of biophysical variables in vegetation.
\newblock \emph{ISPRS Journal of Photogrammetry and Remote Sensing}, 82:\penalty0 83–92, Aug. 2013.
\newblock ISSN 0924-2716.
\newblock \doi{10.1016/j.isprsjprs.2013.04.007}.
\newblock URL \url{http://dx.doi.org/10.1016/j.isprsjprs.2013.04.007}.

\bibitem[Harris et~al.(2020)Harris, Millman, van~der Walt, Gommers, Virtanen, Cournapeau, Wieser, Taylor, Berg, Smith, Kern, Picus, Hoyer, van Kerkwijk, Brett, Haldane, del Río, Wiebe, Peterson, Gérard-Marchant, Sheppard, Reddy, Weckesser, Abbasi, Gohlke, and Oliphant]{harris2020numpy}
C.~R. Harris, K.~J. Millman, S.~J. van~der Walt, R.~Gommers, P.~Virtanen, D.~Cournapeau, E.~Wieser, J.~Taylor, S.~Berg, N.~J. Smith, R.~Kern, M.~Picus, S.~Hoyer, M.~H. van Kerkwijk, M.~Brett, A.~Haldane, J.~F. del Río, M.~Wiebe, P.~Peterson, P.~Gérard-Marchant, K.~Sheppard, T.~Reddy, W.~Weckesser, H.~Abbasi, C.~Gohlke, and T.~E. Oliphant.
\newblock Array programming with numpy.
\newblock \emph{Nature}, 585\penalty0 (7825):\penalty0 357–362, Sept. 2020.
\newblock ISSN 1476-4687.
\newblock \doi{10.1038/s41586-020-2649-2}.
\newblock URL \url{http://dx.doi.org/10.1038/s41586-020-2649-2}.

\bibitem[Hong et~al.(2023)Hong, Zhang, Li, Li, Li, Yao, Yokoya, Li, Ghamisi, Jia, Plaza, Gamba, Benediktsson, and Chanussot]{hong2023spectralgpt}
D.~Hong, B.~Zhang, X.~Li, Y.~Li, C.~Li, J.~Yao, N.~Yokoya, H.~Li, P.~Ghamisi, X.~Jia, A.~Plaza, P.~Gamba, J.~A. Benediktsson, and J.~Chanussot.
\newblock Spectralgpt: Spectral remote sensing foundation model.
\newblock 2023.
\newblock \doi{10.48550/ARXIV.2311.07113}.
\newblock URL \url{https://arxiv.org/abs/2311.07113}.

\bibitem[Hoyer and Hamman(2017)]{hoyer2017xarray}
S.~Hoyer and J.~Hamman.
\newblock xarray: N-d labeled arrays and datasets in python.
\newblock \emph{Journal of Open Research Software}, 5\penalty0 (1):\penalty0 10, Apr. 2017.
\newblock ISSN 2049-9647.
\newblock \doi{10.5334/jors.148}.
\newblock URL \url{http://dx.doi.org/10.5334/jors.148}.

\bibitem[Jakubik et~al.(2023)Jakubik, Roy, Phillips, Fraccaro, Godwin, Zadrozny, Szwarcman, Gomes, Nyirjesy, Edwards, Kimura, Simumba, Chu, Mukkavilli, Lambhate, Das, Bangalore, Oliveira, Muszynski, Ankur, Ramasubramanian, Gurung, Khallaghi, Hanxi, Cecil, Ahmadi, Kordi, Alemohammad, Maskey, Ganti, Weldemariam, and Ramachandran]{jakubik2023prithvi100m}
J.~Jakubik, S.~Roy, C.~E. Phillips, P.~Fraccaro, D.~Godwin, B.~Zadrozny, D.~Szwarcman, C.~Gomes, G.~Nyirjesy, B.~Edwards, D.~Kimura, N.~Simumba, L.~Chu, S.~K. Mukkavilli, D.~Lambhate, K.~Das, R.~Bangalore, D.~Oliveira, M.~Muszynski, K.~Ankur, M.~Ramasubramanian, I.~Gurung, S.~Khallaghi, L.~Hanxi, M.~Cecil, M.~Ahmadi, F.~Kordi, H.~Alemohammad, M.~Maskey, R.~Ganti, K.~Weldemariam, and R.~Ramachandran.
\newblock Foundation models for generalist geospatial artificial intelligence.
\newblock 2023.
\newblock \doi{10.48550/ARXIV.2310.18660}.
\newblock URL \url{https://arxiv.org/abs/2310.18660}.

\bibitem[Knohl et~al.(2003)Knohl, Schulze, Kolle, and Buchmann]{knohl2003hainich}
A.~Knohl, E.-D. Schulze, O.~Kolle, and N.~Buchmann.
\newblock Large carbon uptake by an unmanaged 250-year-old deciduous forest in central germany.
\newblock \emph{Agricultural and Forest Meteorology}, 118\penalty0 (3–4):\penalty0 151–167, Sept. 2003.
\newblock ISSN 0168-1923.
\newblock \doi{10.1016/s0168-1923(03)00115-1}.
\newblock URL \url{http://dx.doi.org/10.1016/S0168-1923(03)00115-1}.

\bibitem[Li and Strahler(1992)]{li1992geo_optical_brdf}
X.~Li and A.~Strahler.
\newblock Geometric-optical bidirectional reflectance modeling of the discrete crown vegetation canopy: effect of crown shape and mutual shadowing.
\newblock \emph{IEEE Transactions on Geoscience and Remote Sensing}, 30\penalty0 (2):\penalty0 276–292, Mar. 1992.
\newblock ISSN 0196-2892.
\newblock \doi{10.1109/36.134078}.
\newblock URL \url{http://dx.doi.org/10.1109/36.134078}.

\bibitem[Lucht et~al.(2000)Lucht, Schaaf, and Strahler]{lucht2000albedo_brdf_model}
W.~Lucht, C.~Schaaf, and A.~Strahler.
\newblock An algorithm for the retrieval of albedo from space using semiempirical brdf models.
\newblock \emph{IEEE Transactions on Geoscience and Remote Sensing}, 38\penalty0 (2):\penalty0 977–998, Mar. 2000.
\newblock ISSN 0196-2892.
\newblock \doi{10.1109/36.841980}.
\newblock URL \url{http://dx.doi.org/10.1109/36.841980}.

\bibitem[Mahecha et~al.(2020)Mahecha, Gans, Brandt, Christiansen, Cornell, Fomferra, Kraemer, Peters, Bodesheim, Camps-Valls, Donges, Dorigo, Estupinan-Suarez, Gutierrez-Velez, Gutwin, Jung, Londoño, Miralles, Papastefanou, and Reichstein]{mahecha2020esdc}
M.~D. Mahecha, F.~Gans, G.~Brandt, R.~Christiansen, S.~E. Cornell, N.~Fomferra, G.~Kraemer, J.~Peters, P.~Bodesheim, G.~Camps-Valls, J.~F. Donges, W.~Dorigo, L.~M. Estupinan-Suarez, V.~H. Gutierrez-Velez, M.~Gutwin, M.~Jung, M.~C. Londoño, D.~G. Miralles, P.~Papastefanou, and M.~Reichstein.
\newblock Earth system data cubes unravel global multivariate dynamics.
\newblock \emph{Earth System Dynamics}, 11\penalty0 (1):\penalty0 201–234, Feb. 2020.
\newblock ISSN 2190-4987.
\newblock \doi{10.5194/esd-11-201-2020}.
\newblock URL \url{http://dx.doi.org/10.5194/esd-11-201-2020}.

\bibitem[Montero et~al.(2023{\natexlab{a}})Montero, Aybar, Mahecha, Martinuzzi, S\"{o}chting, and Wieneke]{montero2023asi}
D.~Montero, C.~Aybar, M.~D. Mahecha, F.~Martinuzzi, M.~S\"{o}chting, and S.~Wieneke.
\newblock A standardized catalogue of spectral indices to advance the use of remote sensing in earth system research.
\newblock \emph{Scientific Data}, 10\penalty0 (1), Apr. 2023{\natexlab{a}}.
\newblock ISSN 2052-4463.
\newblock \doi{10.1038/s41597-023-02096-0}.
\newblock URL \url{http://dx.doi.org/10.1038/s41597-023-02096-0}.

\bibitem[Montero et~al.(2023{\natexlab{b}})Montero, Kraemer, Anghelea, Aybar, Brandt, Camps-Valls, Cremer, Flik, Gans, Habershon, Ji, Kattenborn, Martínez-Ferrer, Martinuzzi, Reinhardt, S\"{o}chting, Teber, and Mahecha]{montero2023datacubes_challenges}
D.~Montero, G.~Kraemer, A.~Anghelea, C.~Aybar, G.~Brandt, G.~Camps-Valls, F.~Cremer, I.~Flik, F.~Gans, S.~Habershon, C.~Ji, T.~Kattenborn, L.~Martínez-Ferrer, F.~Martinuzzi, M.~Reinhardt, M.~S\"{o}chting, K.~Teber, and M.~Mahecha.
\newblock Data cubes for earth system research: Challenges ahead.
\newblock July 2023{\natexlab{b}}.
\newblock \doi{10.31223/x58m2v}.
\newblock URL \url{http://dx.doi.org/10.31223/X58M2V}.

\bibitem[Montero et~al.(2024)Montero, Aybar, Ji, Kraemer, S\"{o}chting, Teber, and Mahecha]{montero2024ondemand}
D.~Montero, C.~Aybar, C.~Ji, G.~Kraemer, M.~S\"{o}chting, K.~Teber, and M.~D. Mahecha.
\newblock On-demand earth system data cubes.
\newblock 2024.
\newblock \doi{10.48550/ARXIV.2404.13105}.
\newblock URL \url{https://doi.org/10.48550/arXiv.2404.13105}.

\bibitem[Pabon-Moreno et~al.(2022)Pabon-Moreno, Migliavacca, Reichstein, and Mahecha]{pabon2022potential_s2_gpp}
D.~E. Pabon-Moreno, M.~Migliavacca, M.~Reichstein, and M.~D. Mahecha.
\newblock On the potential of sentinel-2 for estimating gross primary production.
\newblock \emph{IEEE Transactions on Geoscience and Remote Sensing}, 60:\penalty0 1–12, 2022.
\newblock ISSN 1558-0644.
\newblock \doi{10.1109/tgrs.2022.3152272}.
\newblock URL \url{http://dx.doi.org/10.1109/TGRS.2022.3152272}.

\bibitem[Rocklin(2015)]{rocklin2015dask}
M.~Rocklin.
\newblock Dask: Parallel computation with blocked algorithms and task scheduling.
\newblock In \emph{Proceedings of the Python in Science Conference}, SciPy. SciPy, 2015.
\newblock \doi{10.25080/majora-7b98e3ed-013}.
\newblock URL \url{http://dx.doi.org/10.25080/Majora-7b98e3ed-013}.

\bibitem[Ross(1981)]{ross1981radiation_plant_stands}
J.~Ross.
\newblock \emph{The radiation regime and architecture of plant stands}.
\newblock Springer Netherlands, 1981.
\newblock ISBN 9789400986473.
\newblock \doi{10.1007/978-94-009-8647-3}.
\newblock URL \url{http://dx.doi.org/10.1007/978-94-009-8647-3}.

\bibitem[Roujean et~al.(1992)Roujean, Leroy, and Deschamps]{roujean1992kvol}
J.~Roujean, M.~Leroy, and P.~Deschamps.
\newblock A bidirectional reflectance model of the earth’s surface for the correction of remote sensing data.
\newblock \emph{Journal of Geophysical Research: Atmospheres}, 97\penalty0 (D18):\penalty0 20455–20468, Dec. 1992.
\newblock ISSN 0148-0227.
\newblock \doi{10.1029/92jd01411}.
\newblock URL \url{http://dx.doi.org/10.1029/92JD01411}.

\bibitem[Rouse et~al.(1974)Rouse, Haas, Schell, and Deering]{rouse1974ndvi}
J.~W. Rouse, R.~H. Haas, J.~A. Schell, and D.~W. Deering.
\newblock {Monitoring vegetation systems in the Great Plains with ERTS}.
\newblock Technical report, NASA, 1974.
\newblock URL \url{https://ntrs.nasa.gov/citations/19740022614}.

\bibitem[Roy et~al.(2016)Roy, Zhang, Ju, Gomez-Dans, Lewis, Schaaf, Sun, Li, Huang, and Kovalskyy]{roy2016landsat_nbar}
D.~Roy, H.~Zhang, J.~Ju, J.~Gomez-Dans, P.~Lewis, C.~Schaaf, Q.~Sun, J.~Li, H.~Huang, and V.~Kovalskyy.
\newblock A general method to normalize landsat reflectance data to nadir brdf adjusted reflectance.
\newblock \emph{Remote Sensing of Environment}, 176:\penalty0 255–271, Apr. 2016.
\newblock ISSN 0034-4257.
\newblock \doi{10.1016/j.rse.2016.01.023}.
\newblock URL \url{http://dx.doi.org/10.1016/j.rse.2016.01.023}.

\bibitem[Roy et~al.(2017{\natexlab{a}})Roy, Li, and Zhang]{roy2017rededge}
D.~Roy, Z.~Li, and H.~Zhang.
\newblock Adjustment of sentinel-2 multi-spectral instrument (msi) red-edge band reflectance to nadir brdf adjusted reflectance (nbar) and quantification of red-edge band brdf effects.
\newblock \emph{Remote Sensing}, 9\penalty0 (12):\penalty0 1325, Dec. 2017{\natexlab{a}}.
\newblock ISSN 2072-4292.
\newblock \doi{10.3390/rs9121325}.
\newblock URL \url{http://dx.doi.org/10.3390/rs9121325}.

\bibitem[Roy et~al.(2008)Roy, Ju, Lewis, Schaaf, Gao, Hansen, and Lindquist]{roy2008modis_landsat}
D.~P. Roy, J.~Ju, P.~Lewis, C.~Schaaf, F.~Gao, M.~Hansen, and E.~Lindquist.
\newblock Multi-temporal modis–landsat data fusion for relative radiometric normalization, gap filling, and prediction of landsat data.
\newblock \emph{Remote Sensing of Environment}, 112\penalty0 (6):\penalty0 3112–3130, June 2008.
\newblock ISSN 0034-4257.
\newblock \doi{10.1016/j.rse.2008.03.009}.
\newblock URL \url{http://dx.doi.org/10.1016/j.rse.2008.03.009}.

\bibitem[Roy et~al.(2017{\natexlab{b}})Roy, Li, Zhang, Yan, Huang, and Li]{roy2017sentinel2examination}
D.~P. Roy, J.~Li, H.~K. Zhang, L.~Yan, H.~Huang, and Z.~Li.
\newblock Examination of sentinel-2a multi-spectral instrument (msi) reflectance anisotropy and the suitability of a general method to normalize msi reflectance to nadir brdf adjusted reflectance.
\newblock \emph{Remote Sensing of Environment}, 199:\penalty0 25–38, Sept. 2017{\natexlab{b}}.
\newblock ISSN 0034-4257.
\newblock \doi{10.1016/j.rse.2017.06.019}.
\newblock URL \url{http://dx.doi.org/10.1016/j.rse.2017.06.019}.

\bibitem[S\"{o}chting et~al.(2024)S\"{o}chting, Mahecha, Montero, and Scheuermann]{soechting2024lexcube}
M.~S\"{o}chting, M.~D. Mahecha, D.~Montero, and G.~Scheuermann.
\newblock Lexcube: Interactive visualization of large earth system data cubes.
\newblock \emph{IEEE Computer Graphics and Applications}, 44\penalty0 (1):\penalty0 25–37, Jan. 2024.
\newblock ISSN 1558-1756.
\newblock \doi{10.1109/mcg.2023.3321989}.
\newblock URL \url{http://dx.doi.org/10.1109/MCG.2023.3321989}.

\bibitem[Sumbul et~al.(2019)Sumbul, Charfuelan, Demir, and Markl]{sumbul2019bigearthnet}
G.~Sumbul, M.~Charfuelan, B.~Demir, and V.~Markl.
\newblock Bigearthnet: A large-scale benchmark archive for remote sensing image understanding.
\newblock In \emph{IGARSS 2019 - 2019 IEEE International Geoscience and Remote Sensing Symposium}. IEEE, July 2019.
\newblock \doi{10.1109/igarss.2019.8900532}.
\newblock URL \url{http://dx.doi.org/10.1109/IGARSS.2019.8900532}.

\bibitem[Vaswani et~al.(2017)Vaswani, Shazeer, Parmar, Uszkoreit, Jones, Gomez, Kaiser, and Polosukhin]{vaswani2017transformer}
A.~Vaswani, N.~Shazeer, N.~Parmar, J.~Uszkoreit, L.~Jones, A.~N. Gomez, L.~Kaiser, and I.~Polosukhin.
\newblock Attention is all you need.
\newblock 2017.
\newblock \doi{10.48550/ARXIV.1706.03762}.
\newblock URL \url{https://arxiv.org/abs/1706.03762}.

\bibitem[Walther et~al.(2022)Walther, Besnard, Nelson, El-Madany, Migliavacca, Weber, Carvalhais, Ermida, Br\"{u}mmer, Schrader, Prokushkin, Panov, and Jung]{walther2022fluxneteo}
S.~Walther, S.~Besnard, J.~A. Nelson, T.~S. El-Madany, M.~Migliavacca, U.~Weber, N.~Carvalhais, S.~L. Ermida, C.~Br\"{u}mmer, F.~Schrader, A.~S. Prokushkin, A.~V. Panov, and M.~Jung.
\newblock Technical note: A view from space on global flux towers by modis and landsat: the fluxneteo data set.
\newblock \emph{Biogeosciences}, 19\penalty0 (11):\penalty0 2805–2840, June 2022.
\newblock ISSN 1726-4189.
\newblock \doi{10.5194/bg-19-2805-2022}.
\newblock URL \url{http://dx.doi.org/10.5194/bg-19-2805-2022}.

\bibitem[Wanner et~al.(1995)Wanner, Li, and Strahler]{wanner1995kgeo}
W.~Wanner, X.~Li, and A.~H. Strahler.
\newblock On the derivation of kernels for kernel‐driven models of bidirectional reflectance.
\newblock \emph{Journal of Geophysical Research: Atmospheres}, 100\penalty0 (D10):\penalty0 21077–21089, Oct. 1995.
\newblock ISSN 0148-0227.
\newblock \doi{10.1029/95jd02371}.
\newblock URL \url{http://dx.doi.org/10.1029/95JD02371}.

\bibitem[Yuan et~al.(2022)Yuan, Lin, Liu, Hang, and Zhou]{yuan2022sitsformer}
Y.~Yuan, L.~Lin, Q.~Liu, R.~Hang, and Z.-G. Zhou.
\newblock Sits-former: A pre-trained spatio-spectral-temporal representation model for sentinel-2 time series classification.
\newblock \emph{International Journal of Applied Earth Observation and Geoinformation}, 106:\penalty0 102651, Feb. 2022.
\newblock ISSN 1569-8432.
\newblock \doi{10.1016/j.jag.2021.102651}.
\newblock URL \url{http://dx.doi.org/10.1016/j.jag.2021.102651}.

\end{thebibliography}

\end{document}